%% file: root.tex
\documentclass[letterpaper, 10 pt, conference]{ieeeconf}  % Comment this line out if you need a4paper

\IEEEoverridecommandlockouts                              % This command is only needed if 
\usepackage{graphics} % for pdf, bitmapped graphics files
\usepackage{amsmath} % assumes amsmath package installed
\usepackage{amssymb}  % assumes amsmath package installed
\usepackage{mathtools}
\usepackage{siunitx}
\usepackage{subcaption}
\usepackage{hyperref}
\usepackage{flushend} % aligns references on final page

\title{\LARGE \bf
Evaluating Global Geo-alignment for Precision Learned Autonomous Vehicle Localization using Aerial Data
}

\author{Yi Yang$^{*}$, Xuran Zhao, H. Charles Zhao, Shumin Yuan, Samuel M. Bateman, \\  Tiffany A. Huang, Chris Beall and Will Maddern
\thanks{Authors are with Nuro Inc, Mountain View, CA, USA.}
\thanks{$^{*}$Corresponding author: {\tt\small jayang@nuro.ai}}%
}
\DeclarePairedDelimiter\norm{\lVert}{\rVert}
\DeclareMathOperator*{\argmin}{arg\,min}

\newcommand{\SEii}{\text{SE}(2)}

\newcommand{\SEiii}{\text{SE}(3)}

\newcommand{\seiii}{\mathfrak{se}(3)}

\newcommand{\mapgeo}{\mathbf{M}^g}
\newcommand{\mapcar}{\mathbf{M}^c}
\newcommand{\mapnode}{\mathbf{M}^n}
\newcommand{\submap}{\mathbf{M}_{i, j}}
\newcommand{\submapcar}{\mathbf{M}^c_{i, j}}
\newcommand{\submapgeo}{\mathbf{M}^g_{i, j}}

\newcommand{\Kij}{\mathbf{K}_{i, j}}
\newcommand{\estKij}{\hat{\mathbf{K}}_{i, j}}
\newcommand{\kay}{\mathbf{k}}
\newcommand{\kij}{\mathbf{k}_{i, j}}
\newcommand{\estkij}{\hat{\mathbf{k}}_{i, j}}

\newcommand{\etal}{\textit{et al.\ }}

\begin{document}

\maketitle
\thispagestyle{empty}
\pagestyle{empty}

%%%%%%%%%%%%%%%%%%%%%%%%%%%%%%%%%%%%%%%%%%%%%%%%%%%%%%%%%%%%%%%%%%%%%%%%%%%%%%%%
\begin{abstract}
Recently there has been growing interest in the use of aerial and satellite map data for autonomous vehicles, primarily due to its potential for significant cost reduction and enhanced scalability. 
Despite the advantages, aerial data also comes with challenges such as a sensor-modality gap and a viewpoint difference gap. 
Learned localization methods have shown promise for overcoming these challenges to provide precise metric localization for autonomous vehicles.
Most learned localization methods rely on coarsely aligned ground truth, or implicit consistency-based methods to learn the localization task -- however, in this paper we find that improving the alignment between aerial data and autonomous vehicle sensor data at training time is critical to the performance of a learning-based localization system. 
We compare two data alignment methods using a factor graph framework and, using these methods, we then evaluate the effects of closely aligned ground truth on learned localization accuracy through ablation studies. 
Finally, we evaluate a learned localization system using the data alignment methods on a comprehensive (1600km) autonomous vehicle dataset and demonstrate localization error below 0.3m and 0.5$^{\circ}$ sufficient for autonomous vehicle applications.
\end{abstract}

\input{sec/1_introduction}

\input{sec/2_related_work}

\input{sec/3_methodology}

\input{sec/4_experimental_setup}

\input{sec/5_results}

\input{sec/6_conclusion}


%%%%%%%%%%%%%%%%%%%%%%%%%%%%%%%%%%%%%%%%%%%%%%%%%%%%%%%%%%%%%%%%%%%%%%%%%%%%%%%%
%%%%%%%%%%%%%%%%%%%%%%%%%%%%%%%%%%%%%%%%%%%%%%%%%%%%%%%%%%%%%%%%%%%%%%%%%%%%%%%%
%%%%%%%%%%%%%%%%%%%%%%%%%%%%%%%%%%%%%%%%%%%%%%%%%%%%%%%%%%%%%%%%%%%%%%%%%%%%%%%%
%\section*{APPENDIX}
%\section*{ACKNOWLEDGMENT}
%%%%%%%%%%%%%%%%%%%%%%%%%%%%%%%%%%%%%%%%%%%%%%%%%%%%%%%%%%%%%%%%%%%%%%%%%%%%%%%%
% \begin{thebibliography}{99}
% \end{thebibliography}
\bibliographystyle{IEEEtran}
\bibliography{references}

\end{document}

%% file: sec/1_introduction.tex
\section{Introduction}
Autonomous vehicle localization systems have typically relied on a purpose-built prior map and a fleet of  vehicles to build, maintain, and update the map \cite{Levinson2007Map-BasedEnvironments}, \cite{Wan2018RobustScenes}, limiting the scalability and efficiency of the associated systems. 
In contrast, high quality imagery and 3D elevation data collected from satellite and aerial platforms are increasingly available at relatively low cost, and provide an attractive option to aid or supplant autonomous vehicle localization systems.
However, leveraging aerial data sources to provide ground vehicle localization with sufficient accuracy for autonomous vehicle applications -- typically below 0.3m and 0.5$^{\circ}$ error for urban roads \cite{Reid2019LocalizationVehicles} -- remains a challenge.

Two fundamental obstacles in using aerial data in the vehicle localization task are a sensor modality gap and a view-point difference gap.
Recent work such as \cite{Tang2021Self-supervisedLocalization, Noda2011VehicleImage, Shi2022BeyondImage, Sarlin2023Orienternet:Matching, Jin2024BEVRender:Environment} have taken steps towards resolving these challenges by learning view-invariant, cross-modal features across disparate sensing modalities, resulting in systems capable of achieving meter-level localization accuracy at a global scale \cite{Tang2021Self-supervisedLocalization,Shi2022BeyondImage,Jin2024BEVRender:Environment}. 
However, this still leaves a significant performance gap compared to the sub-meter accuracy achieved by traditional localization systems found in typical autonomous vehicles \cite{Reid2019LocalizationVehicles}. 

\begin{figure}[t!]
        \includegraphics[width=\linewidth]{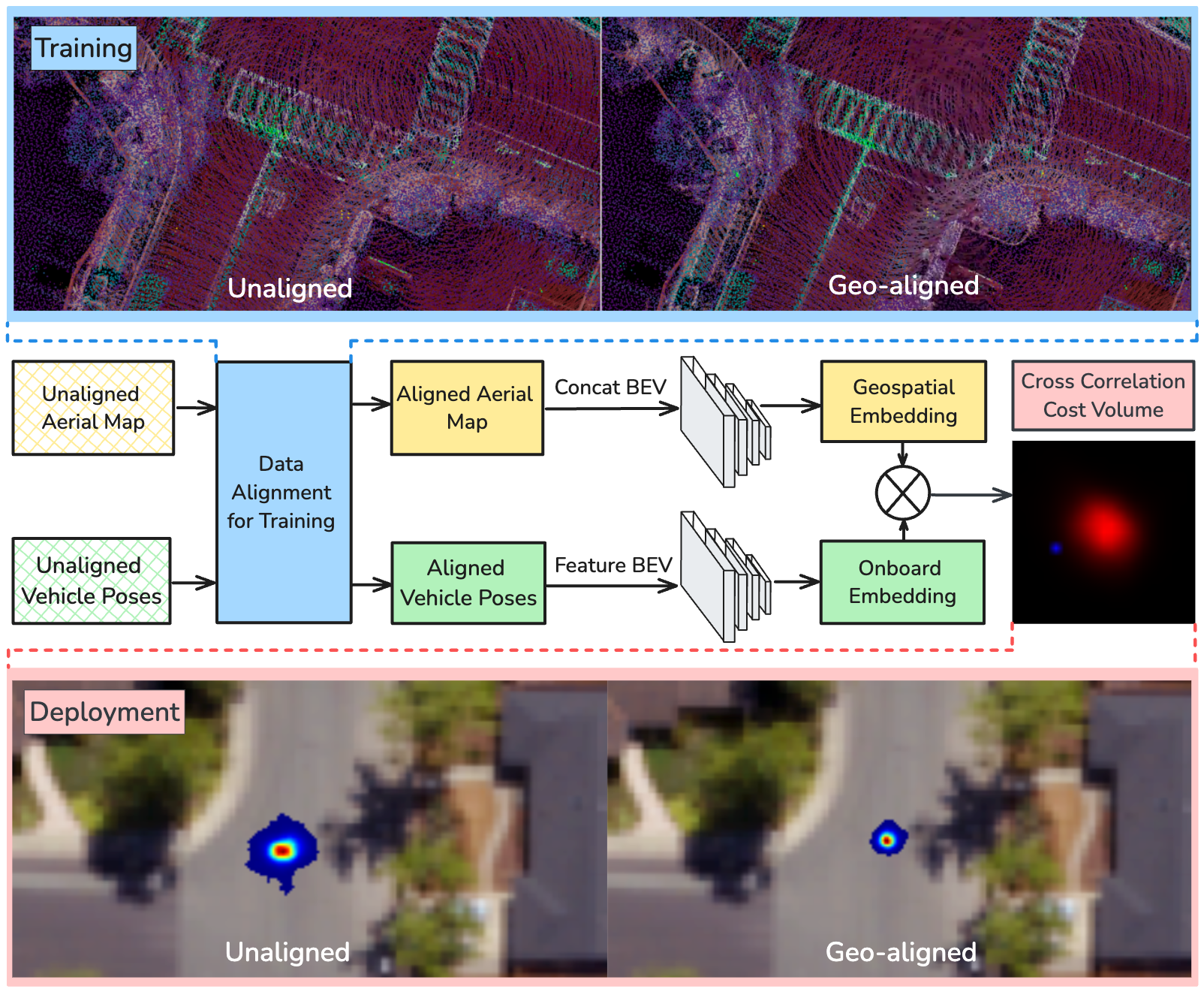}
        \label{fig:overview}

    \caption{
    We show that explicit alignment between vehicle LiDAR and aerial DSM at training time (top) is key to unlocking sub-meter-accurate learned global localization using aerial imagery on urban roads (bottom).
    In this paper we compare approaches for map data geo-alignment (blue box) at training time for a learning-based autonomous vehicle localization system.
    The resulting localization solutions are significantly improved at run-time to an accuracy sufficient for autonomous vehicle applications.
    }
    \label{fig:cover_figure}
    \vspace{-10pt}
\end{figure}

In this work, we identify that a key to unlock sub-meter vehicle localization accuracy using aerial data is the precise choice of data alignment method between aerial data and the vehicle sensor data used for training a learning-based system. 
Although post-processed GNSS should in-theory provide accurate localization of the vehicle trajectory with respect to the geospatial map, alignment errors still persist due to local atmospheric conditions and variations in aerial data processing stages. 
Previous work \cite{Shi2022BeyondImage} has identified and confirmed this alignment deficiency in existing autonomous vehicle datasets such as KITTI \cite{Geiger2013VisionDataset} and Ford AV \cite{Agarwal2020FordDataset} when combined with aerial data, despite the presence of post-processed GNSS ground truth provided with these datasets. 
With the goal of attaining global sub-meter localization accuracy, rectifying the misaligned geometry before performing supervised learning for localization becomes critical (Fig. \ref{fig:cover_figure}); it removes a significant source of aleatoric uncertainty from the data and simplifies the training objective, allowing the model to focus solely on the cross-modal domain gap.

To this end, in this paper we present two broad categories of methods to align aerial and autonomous vehicle data before using the aligned data to train a learning-based localization system. 
We present a comprehensive experimental evaluation using a large dataset collected using autonomous vehicle data aligned to public domain digital surface model (DSM) data and RGB aerial imagery. 
We include ablation studies to determine the impact of alignment choice and available modalities on global localization accuracy as well as local consistency. 
Finally, we demonstrate that, by leveraging the data alignment process at training time, a learning-based localization system can achieve global sub-meter accuracy sufficient for autonomous vehicle applications using aerial data.

\begin{figure*}[!thbp]
    \centering
    \begin{subfigure}{\linewidth}
        \includegraphics[width=\linewidth]{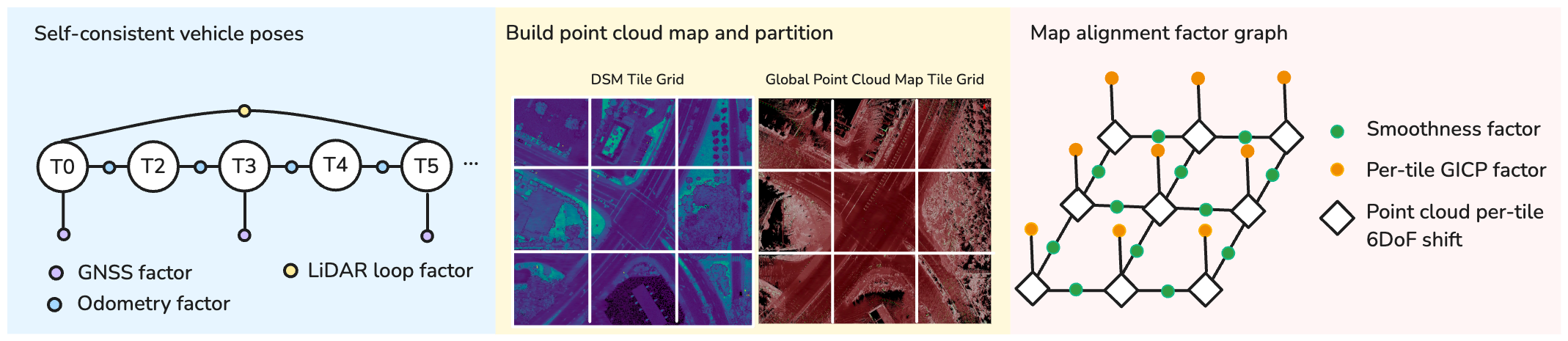}
        \caption{Align aerial map to vehicle data (map-to-vehicle).}
        \label{fig:align_map_to_run}
    \end{subfigure}
    \hfill
    \begin{subfigure}{\linewidth}
        \includegraphics[width=\linewidth]{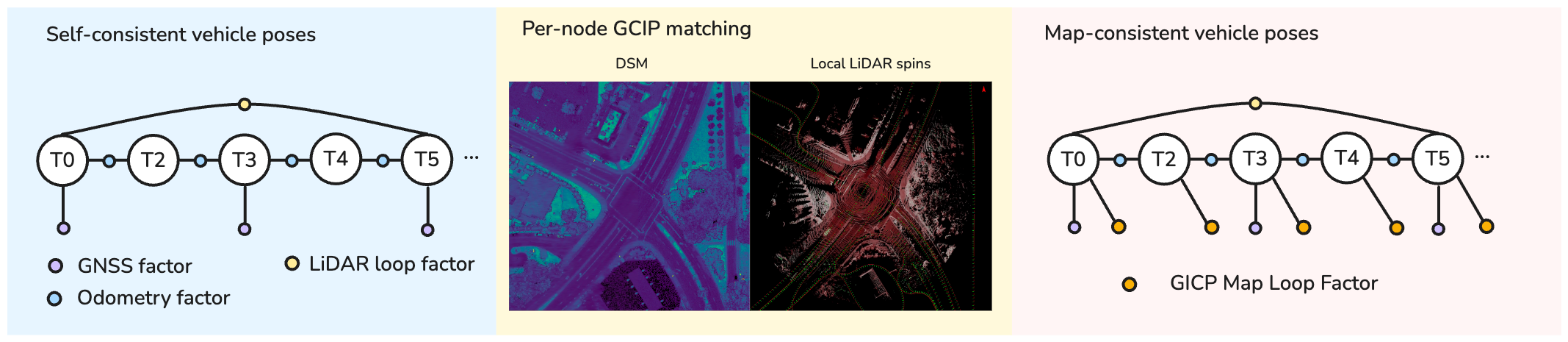}
        \caption{Align vehicle data to aerial map (vehicle-to-map).}
        \label{fig:align_run_to_map}
    \end{subfigure}
    \caption{
    We compare two classes of alignment approach, where (a) shows the workflow to align aerial map data to vehicle data and (b) shows the workflow to align vehicle data to aerial map data.}
    \label{fig:alignment_methods}
    \vspace{-10pt}
\end{figure*}

%% file: sec/2_related_work.tex
\section{Related Work}
\label{sec:related-work}

Localization using aerial or satellite imagery has significant challenges, including perspective and illumination differences and occlusion. 
Historically, geometric feature detection techniques were explored to overcome them \cite{NormanMattern2011VehicleImages, MayankBansal2011Geo-LocalizationDatabases}, as well as sparse feature-point matching \cite{Noda2011VehicleImage}. 
Moreover, autonomous vehicles usually leverage LiDAR sensors in addition to cameras to overcome poor illumination and extreme weather conditions, and thus matching LiDAR grid maps against aerial imagery has gained more and more focus. 
However, these non-learned approaches often fail to capture complex relationships between data modalities and require significant tuning, compromising system reliability \cite{Vora2020AerialVehicles, LucasdePaulaVeronese2015Re-emissionEnvironments}.

\subsection{Map-to-Vehicle Alignment}
\label{sec:map-based-alignment-references}
The problem of registering aerial data to the street-level map collected from vehicles can be regarded as a submap registration problem. 
\cite{OlusholaAbayowa2015AutomaticModels} proposed registering the aerial imagery-derived DSM to ground point cloud by coarsely matching the geometric invariant features such as SIFT \cite{Lowe2004DistinctiveKeypoints}, and later refine the registration using the Iterative Closest Point (ICP) \cite{Besl1992MethodShapes} method. 
\cite{Gawel20173DEstimation} studied the submap registration problem in the context of matching map data between a ground robot and a UAV, evaluating 3D feature descriptors such as Fast Point Feature Histogram \cite{Rusu2009FastRegistration} and Fast Global Registration \cite{Zhou2016FastRegistration} for registering  ground submaps to aerial global maps under different perspective views. 
\cite{Habbecke2010AutomaticMaps} registered aerial imagery  and LiDAR data to a ground vector map representing the building contours by leveraging intrinsic geometric properties of vanishing lines. 
\cite{Vanmiddlesworth2015MappingSubmaps} leveraged smoothness constraints to register multiple submaps for underwater environments.
These geometric constraints are later incorporated as costs in a bundle-adjustment-like batch optimization problem. These studies form the foundation of our map-to-vehicle method.

\subsection{Vehicle-to-Map Alignment}
\label{sec:vehicle-based-alignment-references}
Aligning the vehicle to an external geospatial map is similar to the process of drift-free global localization. 
The process is used to achieve an as-accurate-as-possible localization estimation for the vehicle such that the resulting pose can be used as the ground truth label during the learning phase. 
\cite{Carle2010Long-rangeMaps} studied the problem of localizing a robotic rover equipped with a LiDAR to a high resolution DSM by detecting the topographic peaks using morphological dilation. 
\cite{Kummerle2011LargeInformation} used aerial imagery as prior to loop close to the ground robot LiDAR by extracting the line features. 
\cite{Vora2020AerialVehicles} utilizes Normalized Mutual Information as the key metric to match a LiDAR intensity map against aerial imagery. Similarly, \cite{Patel2020VisualUAVs} proposed using mutual information to localize a UAV against satellite imagery taken from Google Maps. 
\cite{Levinson2007Map-BasedEnvironments} employed phase correlation to localize using a prior-build LiDAR intensity map. 
Given that these methods rely on handcrafted features and specific physical or geometric properties, they have limited success in bridging the sensor modality gap and generalize poorly to different sensor or map types. 

\subsection{Learning-Based Localization}
To overcome the modality gap, recent work has investigated using deep learning to improve geospatial localization performance. 
These approaches can broadly be categorized as addressing either the coarse-grained \textit{geo-localization} task, or the fine-grained \textit{metric localization} task where it is assumed a coarse-grained initial estimate is provided (e.g. from GNSS). 
Deep learned geo-localization approaches (\cite{Downes2022City-wideAgent, Shi2020WhereMatching, Liu2019LendingGeo-localization, Shi2019Spatial-AwareGeo-Localization, Samano2019YouImages, Regmi2019BridgingMatching, Hu2018CVM-Net:Geo-Localization, Vo2016LocalizingImagery}) are generally based on metric learning, where onboard observations and offboard map tiles are encoded into a common latent vector space and similarity is computed to perform retrieval. 
We will focus on metric localization, where the goal is to refine an initial noisy localization estimate.

Bârsan \etal \cite{Barsan2018LearningMap} proposes a learned metric localization approach for localizing onboard LiDAR data against a high definition LiDAR intensity map. 
The LiDAR intensity map is passed through a CNN to produce a 2D map embedding, and a cross-correlation operation is used to localize the vehicle embedding at run-time. 

There has also been much research into using deep learning for cross-modality localization, e.g. localizing ground-level camera images against top-down aerial imagery (\cite{Xia2024ConvolutionalEstimation, Zhang2024IncreasingRegistration, Jin2024BEVRender:Environment, Shi2023AccurateMatching, Shi2022BeyondImage, Zhu2021VIGOR:Retrieval, Chebrolu2019RobotFields, Wang2017FLAG:Ground}), or localizing radar observations against aerial imagery data (\cite{Tang2020RSL-Net:Ground, Tang2021Self-supervisedLocalization}). 
OrienterNet \cite{Sarlin2023Orienternet:Matching} localizes single street-view camera images against 2D navigation maps by first lifting the camera images into BEV, and then localizing in BEV similar to \cite{Barsan2018LearningMap}. 
\cite{Shi2022BeyondImage}, which localizes street-view camera images against aerial imagery, proposes a different approach where they project the aerial imagery into perspective view using a ground plane estimate and then perform alignment in perspective view instead of BEV. 
BEVRender \cite{Jin2024BEVRender:Environment} avoids the need for a separate map encoder, and thus the need to store potentially expensive map embeddings, by introducing a render head that converts from the latent onboard embedding image into an RGB image that can be directly aligned with aerial imagery.

In this work, we adopt a cross-modality, correlation-based learned localization model similar to that of \cite{Barsan2018LearningMap} and \cite{Sarlin2023Orienternet:Matching}.

%% file: sec/3_methodology.tex
\section{Methodology}

We outline a baseline method with two major components: \textit{data alignment} and a \textit{learning-based localization system}. 
Broadly following the two approaches for data alignment covered in Sec. \ref{sec:related-work}, we present two different methods: \textit{map-to-vehicle} which aligns the aerial data to the ground data collected from a vehicle, and \textit{vehicle-to-map} which aligns vehicle data to the aerial map, as shown in Fig. \ref{fig:alignment_methods}. 
Then, we present a baseline learning-based localization system inspired by \cite{Barsan2018LearningMap} that leverages the aligned map and vehicle data to learn a \textit{geospatial embedding} and an \textit{onboard embedding}, each of which contains salient geometric features to enable precise cross-correlation matching for vehicle localization. 
For alignment approaches we assume use of aerial data containing a digital surface model (DSM) that represents a 2.5D/3D surface elevation of the environment, and aligned RGB color imagery. 

\begin{figure*}[ht!]
    \centering
    \includegraphics[width=1.0\textwidth]{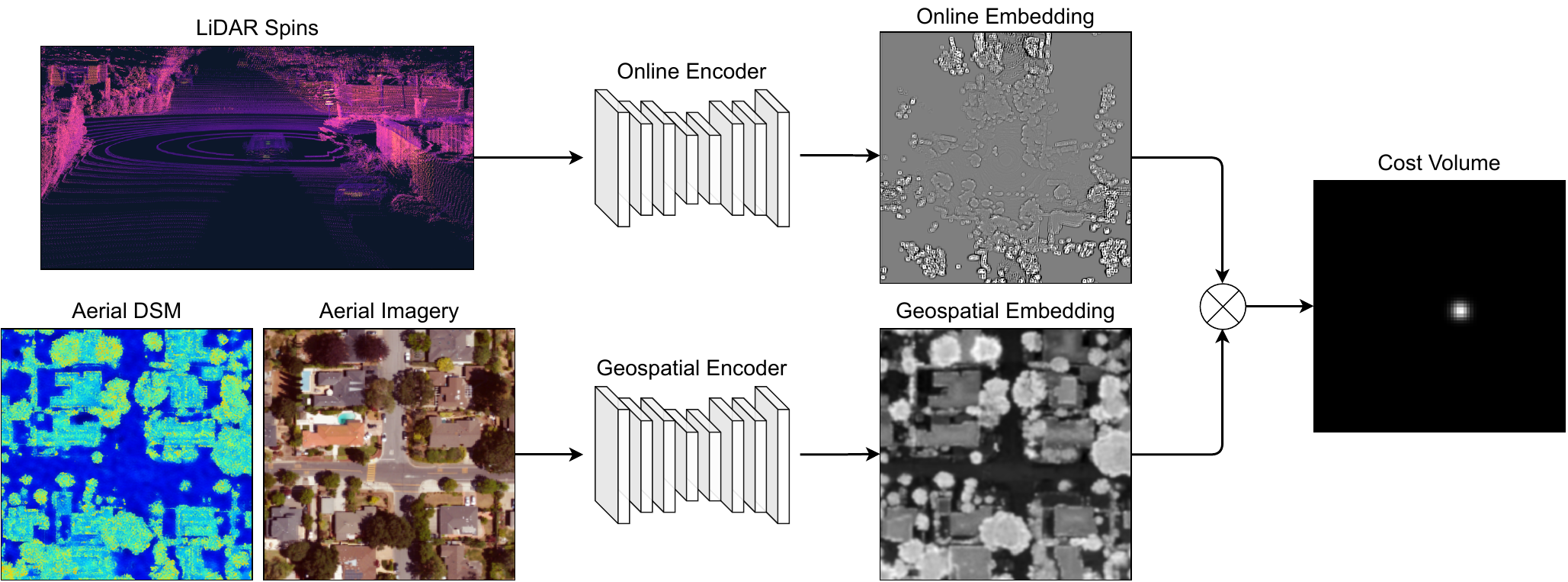}
    \caption{Localization model architecture inspired by \cite{Barsan2018LearningMap, Sarlin2023Orienternet:Matching}. The model consists of two components, an online encoder (top) which consumes onboard LiDAR spins, and a geospatial encoder (bottom) which consumes aerial DSM and/or imagery. The embedding images produced by these two components are aligned with each other by computing the cross correlation over a search window of possible $x$, $y$, and $\theta$ offsets. 
    }
    \label{fig:model_arch}
    \vspace{-10pt}
\end{figure*}

\subsection{Map-to-vehicle alignment}

Following approaches in Sec.\ \ref{sec:map-based-alignment-references}, we start with a \textit{self-consistent factor graph} representing the optimized poses of the data collection vehicle, as shown in the left of Fig. \ref{fig:align_map_to_run}. 
The set of nodes $\mathbf{T}_k \in \SEiii$ represent the optimized vehicle pose at time $k$, with factors containing GNSS, odometry, and LiDAR loop closure measurements. 
The purpose of incorporating GNSS factors is to ensure the self-consistent factor graph is anchored with a good initial estimate such that the subsequent map-matching problem is well-defined.

From the self-consistent factor graph with optimized and fixed vehicle poses, we build a point cloud map $\mapcar$ by projecting the LiDAR points obtained at all times $k$ using the estimated $\mathbf{T}_k$. 
We partition both the point cloud map $\mapcar$ and the DSM channel of the aerial map $\mapgeo$ into a set of tile grids. 
Each tile in the grid is represented by $\submap$, where $i, j$ are the corresponding row and column index. 
Next, we calculate the best possible transformation $\estKij \in \SEiii$ between the tile $\submapgeo$ and the tile $\submapcar$ using generalized Iterative Closest Point (GICP) \cite{Segal2009Generalized-ICP}. 

We create a \textit{map alignment factor graph} where each node represents the transformation $\Kij \in \SEiii$ between the tile $\submapgeo$ and the tile $\submapcar$. 
The set of GICP measurements $\estKij$ is incorporated into the map alignment factor graph as unary factors for each node. Inspired by \cite{Vanmiddlesworth2015MappingSubmaps}, we also use a smoothness factor $\mathbb{I}$, the $\SEiii$ group identity, to preserve smoothness at the tile boundaries. Let $\mathbf{k}_{i, j} \in \mathbb{R}^6 $ be the corresponding Lie algebra of the group element $\Kij$. 
We solve the following optimization problem to obtain a globally-consistent set of transformations to align the aerial data to the vehicle data. 

\begin{align}
    \begin{split}
    \mathcal{K^\ast} = \argmin_\mathcal{K} & \sum_{i=0, j=0}^{N_i, N_j} \norm{ \kij - \estkij }_{\Sigma_{\mathbf{K}}} + \\& \sum_{i=1, j=0}^{N_i, N_j}\norm{\mathbf{s}_{i}}_{\Sigma_{\mathbf{s}}} + \sum_{i=0, j=1}^{N_i, N_j}\norm{\mathbf{s}_{j}}_{\Sigma_{\mathbf{s}}},
    \end{split}
\end{align}

\begin{align}
    \mathbf{s}_{i} = \kij - \kay_{i-1,j},\,\,\,
    \mathbf{s}_{j} = \kij - \kay_{i,j-1}.
\end{align}

\noindent $N_i$ and $N_j$ represent the number of tiles in the horizontal and vertical directions. 
$\mathbf{s}_{i}$ and $\mathbf{s}_{j}$ represent the residuals from the smoothness factor, and $\Sigma_\mathbf{K}$ and $\Sigma_\mathbf{S}$ represent the covariance of the GICP matching and smoothness constraints respectively. 
After optimization, we apply the resulting set of transformations to the $\mapgeo$ of every tile of the grid and produce a new geographic map $\tilde{\mathbf{M}}^g$ that is now aligned to the vehicle data.

\subsection{Vehicle-to-map alignment}
\label{sec:align-car-to-map}
As in the previous section, we start out with a self-consistent factor graph of vehicle poses. For each vehicle pose node $\mathbf{T}_k$ at timestamp $k$, we extend the timestamp by $\tau$ seconds to construct a time interval within which a set of LiDAR spins is collected. Each LiDAR spin corresponds to a vehicle pose $\mathbf{T}_{\tau_i}$. We use this set of vehicle poses to accumulate the LiDAR spins within the time interval, and we denote this set of LiDAR points as submap $\mapnode_{k}$, as shown in Fig. \ref{fig:align_run_to_map}.

For every node in the factor graph, we calculate a 6-DoF transformation $\hat{\mathbf{T}}_k$ between the accumulated LiDAR map $\mapnode_k$ and the geo map $\mapgeo$ using GICP. The set of transformations is incorporated back into the self-consistent factor graph for each vehicle pose as a unary factor. Let $\mathbf{t}_{i} \in \seiii $ be the corresponding Lie algebra of the group element $\mathbf{T}_{i}$ at the $i$-th node, we have:
\begin{equation}
\mathcal{T}^\ast = \argmin_{\mathcal{T}} \sum_{i = 1}^{N} \norm{ \mathbf{R}_{\mathbf{t}_{i}}}_{{\Sigma}_{\mathbf{R}_{i}}} + \sum_{i = 1}^{N} \norm{\mathbf{t}_{i} - \hat{\mathbf{t}}_i}_{\Sigma_{\hat{\mathbf{T}}_i}}
\end{equation}
where $\mathbf{R}_\mathbf{t}$ represents the residual of all measurements excluding the GICP map factors within the self-consistent factor graph, and $\Sigma_{\mathbf{R}_i}$ and $\Sigma_{\hat{\mathbf{T}}_i}$ are the covariances from all measurements of the self-consistent factor graph and GICP map factors.  
The factor graph is re-optimized to obtain a \textit{map consistent factor graph}, representing the optimized poses of the vehicle trajectory in the geospatial map $\mapgeo$ frame. The name \textit{map consistent} is used to indicate that the vehicle is now localized with respect to the aerial map with the newly added map loop factors.

\subsection{Precision learned localization}
\label{sec:learned-localization}

To evaluate the different alignment approaches presented in an autonomous vehicle context, we adopt a learning-based localization approach inspired by \cite{Barsan2018LearningMap}. 
The system diagram is shown in Fig. \ref{fig:model_arch}. 
The first component encodes a set of accumulated onboard LiDAR spins into a 2D ``online embedding." 
The accumulated LiDAR spins are generated following a similar procedure as outlined in Sec \ref{sec:align-car-to-map}. 
For the map-to-vehicle alignment approach, the accumulated LiDAR spins are obtained from the self-consistent factor graph. 
For the vehicle-to-map alignment method, the accumulated LiDAR spins are obtained from the map-consistent factor graph. 
The second component is a fully convolutional geospatial feature encoder network that encodes DSM and/or RGB imagery channels into a single 2D ``geospatial embedding." 
Note that while DSM is required for the alignment stage to generate training data, it is not required to generate the embedding.

The translation offset $x, y$ is recovered using a cross-correlation operation between the online embedding and the geospatial embedding to generate a cost volume, and a Gaussian is fitted to the cost volume for sub-pixel accuracy. 
The orientation offset $\theta$ is solved similarly using a polar image representation.

During training, we use the 6-DoF vehicle poses $\mathbf{T}$ obtained from the data alignment (either the self-consistent factor graph or the map-consistent factor graph) as the training label. 
We define a localization estimate as follows:
\begin{align}
\mathbf{x} = [x, y, \theta] \in \SEii
\end{align}
\noindent where $\mathbf{x}$ is the localization estimate in local frame, with $x$ pointing forward longitudinally, $y$ pointing to the left laterally, and $\theta$ representing yaw angle. 
We use an L1 loss for the Gaussian of the cost volume during training:

\begin{equation}
\label{eq:loc_loss}
\mathcal{L} = \sum_{i\in x, y, \theta} \norm{ \mu_i - \hat{\mu}_i}_{1} + \alpha_i\norm{\sigma_i - \hat{\sigma}_i}_{1}
\end{equation}
\noindent where $\hat{\mathbf{x}}_i \sim \mathcal{N}(\hat{\mu}_i,\,\hat{\sigma}_i)$ is the estimated pose from localization, and $\mathbf{x}_i \sim \mathcal{N}(\mu_i,\,\sigma_i)$ is the ground truth pose. 
The localization system is trained with each ground truth pose randomly perturbed on the $\SEii$ manifold to produce a new pose $\tilde{\mathbf{x}}_i$. 
The new pose is used as the initialization point for cross-correlation, and the offline and the online CNNs learn the features that minimize the cross-correlation cost. 
Since the covariance loss tends to be larger than the mean loss, the sum of two losses are weighted by $\alpha$, which is a value selected empirically.

Despite the data modality gap between the aerial map data and the sensor data, this formulation allows the network to learn the most salient features within each modality as it strives to minimize the cross-correlation costs.

%% file: sec/4_experimental_setup.tex
\section{Experimental Setup}

\begin{figure}[t!]
\vspace{2pt}
\centering
\includegraphics[width=0.48\textwidth]{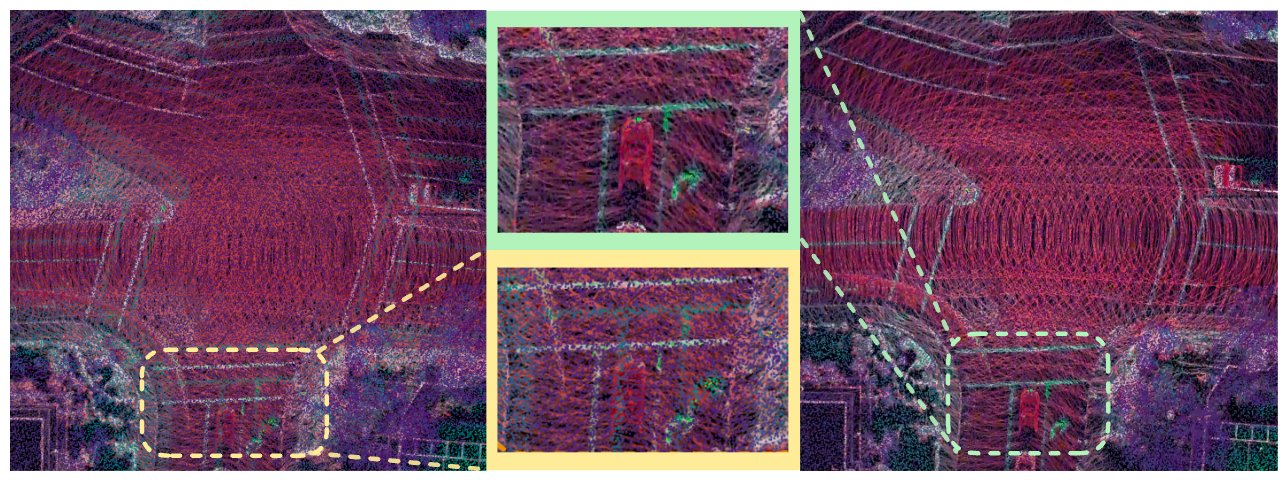}

\caption{Overlay images of USGS DSM data and the vehicle LiDAR map with (left) no alignment and (right) vehicle-to-map alignment. Vehicle LiDAR data is shown in the white-red color, and USGS DSM data is shown in the green-blue color. Note that the LiDAR artifacts in the highlighted region are resolved through the alignment process.}
\label{fig:alignment_eval_comp}
\vspace{-10pt}
\end{figure}

Existing datasets for evaluating autonomous vehicle localization \cite{Geiger2013VisionDataset, Agarwal2020FordDataset, Maddern2015LeveragingCities} either do not contain aerial data or are not sufficiently large for training learning-based localization models.
We present evaluation results using a large-scale vehicle dataset and public United States Geological Survey (USGS) / United States Department of Agriculture (USDA) aerial data in a 147.3 sq km area in the San Francisco Bay Area, USA. 

\subsection{Vehicle Data}
The on-road dataset was collected using an autonomous vehicle equipped with GNSS, Hesai40p LiDAR and IMU traversing urban roads in the San Francisco Bay Area through multiple years under different time and weather conditions, yielding more than $1600\text{km}$ of data.

\subsection{Aerial Data}
We used DSM point cloud data collected from the 3D Elevation Program \cite{U.S.GeologicalSurveyUSGSProgram}, and RGB imagery collected from the National Agricultural Imagery Program \cite{U.S.DepartmentofAgricultureNationalNAIP}. 
We applied a $90/10$ split based on geographic coverage for training and evaluation sets. 
The ground sampling distance of USDA aerial images is $0.6\text{m}$, and the images were upscaled to $0.2\text{m}$ before alignment using a simple pixel-wise linear interpolation.
The tile grids for the map-to-vehicle alignment approach are $32 \times 32$ meters for both aerial and vehicle data.
Example alignment results are shown in Fig. \ref{fig:alignment_eval_comp}.

\subsection{Training Details}
We randomly sampled $640\text{k}$ training examples from our dataset, then accumulated and rasterized 5 LiDAR spins across $0.5\text{s}$ for each example. 
Random perturbation sampled uniformly within $\pm \SI{3}{\deg}, \pm \SI{1}m$ was applied on the ground truth for each example in both training and testing, and used an L1 loss supervising the offset from the ground truth, as defined in equation \ref{eq:loc_loss}. 
We also included a weighted covariance loss encouraging higher confidence from the estimation. 
A higher weight at $\alpha = 0.2$ was applied for yaw covariance, and a lower weight at $\alpha = 0.04$ was applied for covariance along x and y.
We compare the trained models on the evaluation dataset using global RMS, median (P50) and 99th percentile (P99) localization errors.

%% file: sec/5_results.tex
\section{Results}

\subsection{Global Localization}

\begin{figure}[t!]
\vspace{4pt}
\centering
\begin{subfigure}{0.31\columnwidth}
    \includegraphics[width=\textwidth]{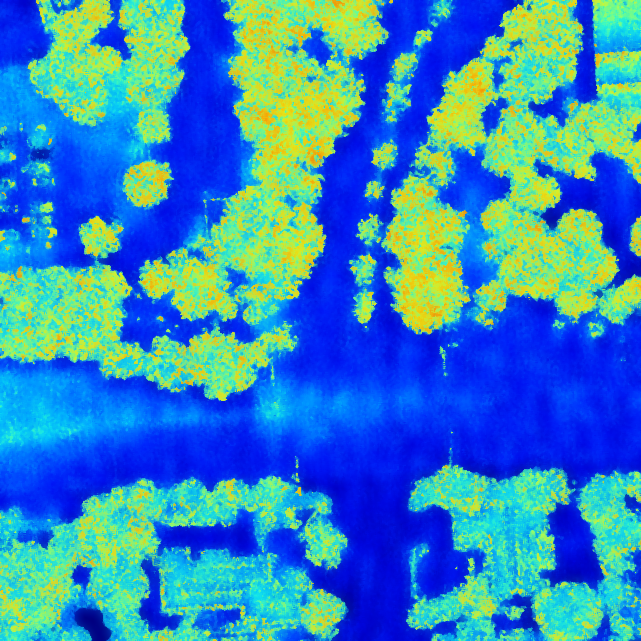}
\end{subfigure}
\hfill
\begin{subfigure}{0.31\columnwidth}
    \includegraphics[width=\textwidth]{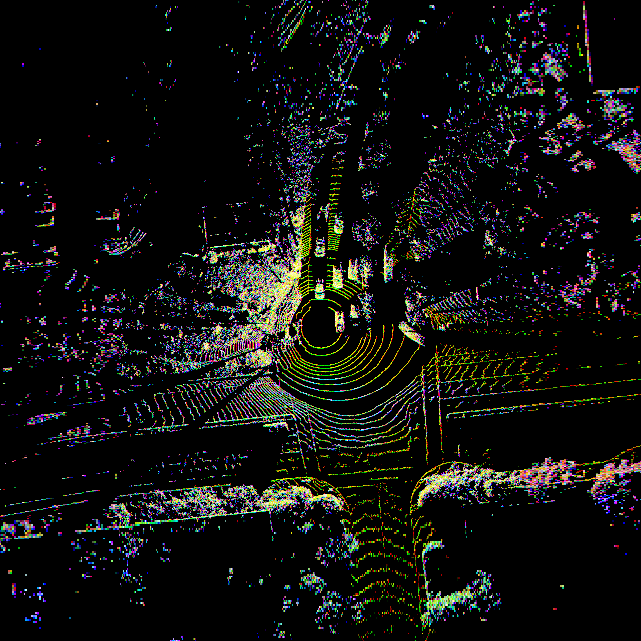}
\end{subfigure}
\hfill
\begin{subfigure}{0.31\columnwidth}
    \includegraphics[width=\textwidth]{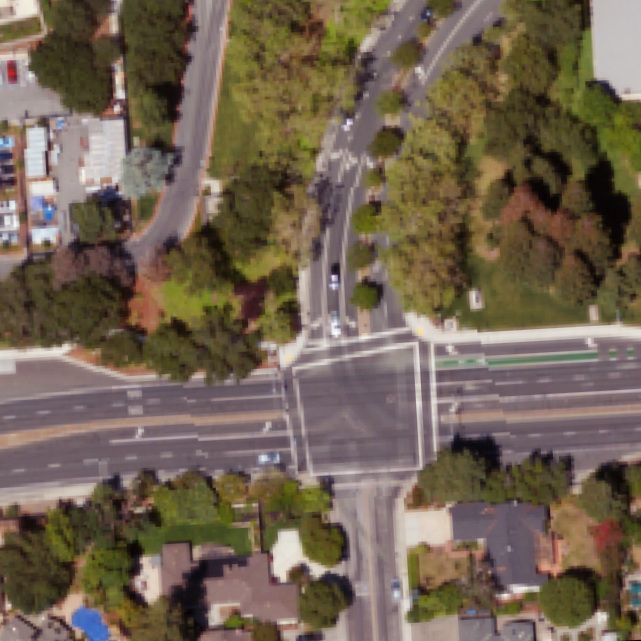}
\end{subfigure}

\vspace{2pt}

\begin{subfigure}{0.31\columnwidth}
    \includegraphics[width=\textwidth]{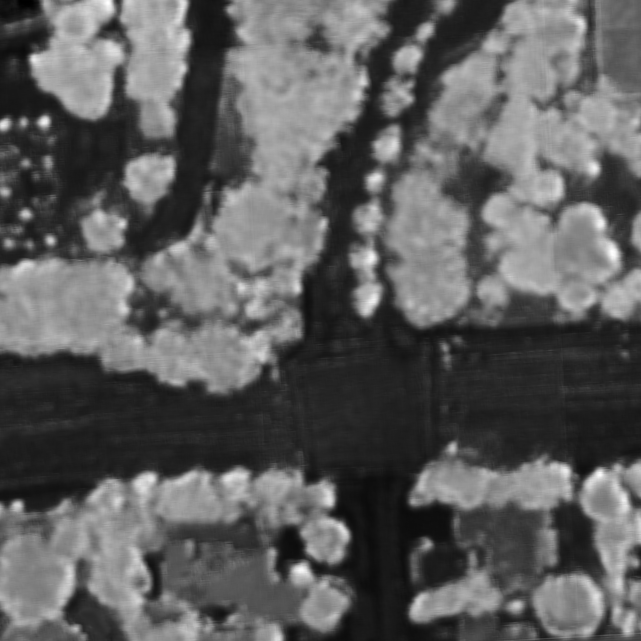}
\end{subfigure}
\hfill
\begin{subfigure}{0.31\columnwidth}
    \includegraphics[width=\textwidth]{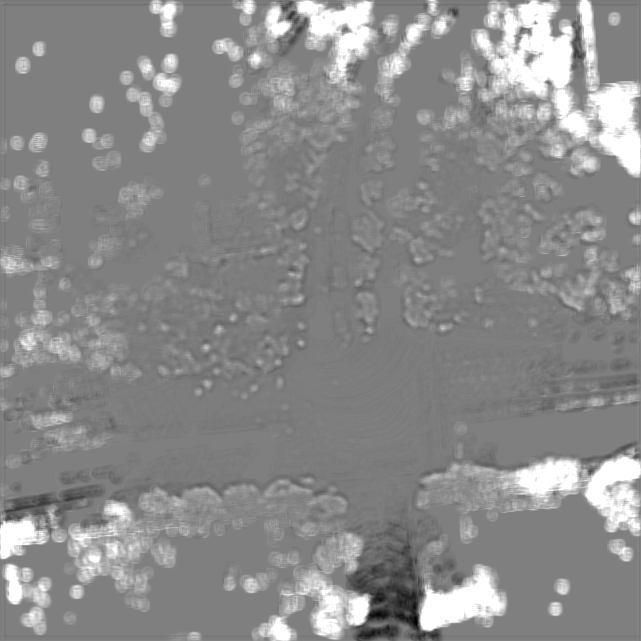}
\end{subfigure}
\hfill
\begin{subfigure}{0.31\columnwidth}
    \includegraphics[width=\textwidth]{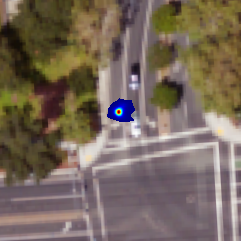}
\end{subfigure}

\vspace{2pt}

\begin{subfigure}{0.31\columnwidth}
    \includegraphics[width=\textwidth]{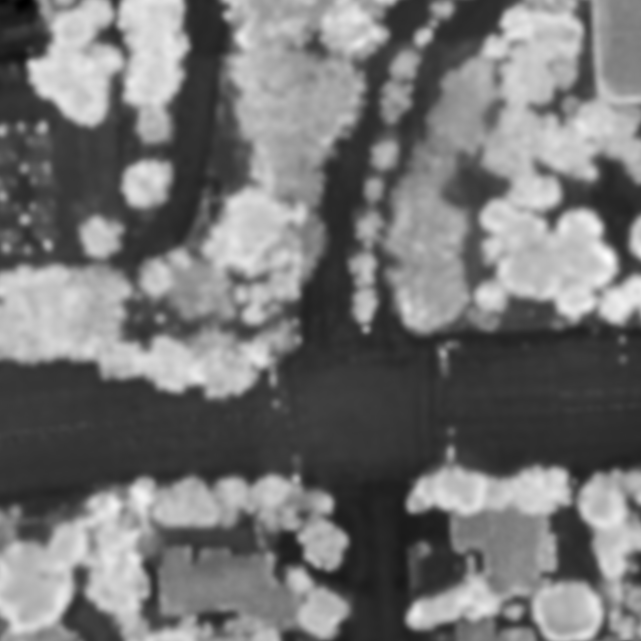}
\end{subfigure}
\hfill
\begin{subfigure}{0.31\columnwidth}
    \includegraphics[width=\textwidth]{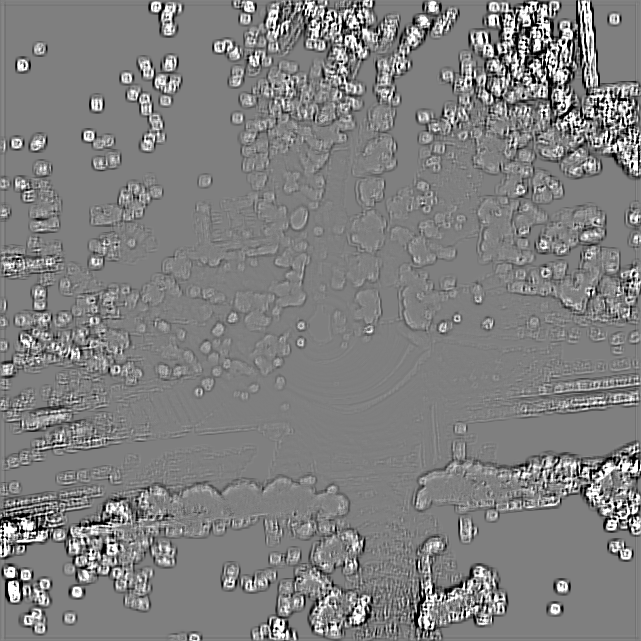}
\end{subfigure}
\hfill
\begin{subfigure}{0.31\columnwidth}
    \includegraphics[width=\linewidth]{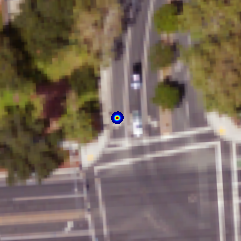}
\end{subfigure}

\caption{
Localization results from models trained using no alignment vs vehicle-to-map alignment. 
The first row from left to right: USGS DSM, online LiDAR, USDA aerial imagery; subsequent rows show models trained without alignment and by vehicle-to-map alignment respectively. 
Note the improved contrast of the online embedding (center) and reduced uncertainty (right) with improved training alignment.
}
\label{fig:map alignment on localization results}
\vspace{-10pt}
\end{figure}

\begin{table*}[htbp]
\label{table_localization_different_alignment}
\vspace{2pt}
\caption{Metric localization results with different alignment approaches and aerial data modalities.}
\begin{center}
\begin{tabular}{c|c|c|c|c|c|c|c|c|c|c}
\hline
\textbf{Geospatial} & \textbf{Aerial} & \multicolumn{3}{|c|}{\textbf{RMS}} & \multicolumn{3}{|c|}{\textbf{P50}} & \multicolumn{3}{|c}{\textbf{P99} } \\
\cline{3-11} 
\textbf{Alignment} & \textbf{Data Modality} & \textbf{\textit{lat (m)}}& \textbf{\textit{lng (m)}}& \textbf{\textit{yaw (deg)}} & \textbf{\textit{lat (m)}}& \textbf{\textit{lng (m)}}& \textbf{\textit{yaw (deg)}} & \textbf{\textit{lat (m)}}& \textbf{\textit{lng (m)}}& \textbf{\textit{yaw (deg)}} \\
\hline
No Alignment & Imagery Only & 0.885 & 0.877 & 0.900 & 0.610 & 0.590 & 0.442 & 2.426 & 2.578 & 2.905\\
No Alignment & DSM Only & 0.988 & 0.893 & 0.828 & 0.748 & 0.697 & 0.371 & 2.433 & 2.455 & 2.728 \\
No Alignment & Imagery + DSM & 0.999 & 0.916 & 0.898 & 0.788 & 0.717 & 0.412 & 2.403 & 2.470 & 2.881\\
\hline
Map to Vehicle & Imagery Only & 1.040 & 1.035 & 1.042 &  0.638 & 0.789 & 0.532 & 2.787 & 2.588 & 3.292 \\
Map to Vehicle & DSM Only & 0.299 & 0.274 & 0.517 & 0.163 & 0.161 & 0.230 & 0.996 & 0.803 & 1.658 \\
Map to Vehicle & Imagery + DSM & 0.273 & 0.247 & 0.462 & 0.161 & 0.160 & 0.226 & 0.782 & 0.741 & 1.453 \\
\hline
Vehicle to Map & Imagery Only & 0.484 & 0.592 & 0.608 & 0.215 & 0.232 & 0.319 & 1.607 & 2.023 & 2.092\\
Vehicle to Map & DSM Only & 0.187 & 0.208 & 0.387 & 0.125 & \textbf{0.134} & 0.187 & 0.471 & 0.533 & \textbf{1.345}\\
Vehicle to Map & Imagery + DSM & \textbf{0.187} & \textbf{0.206} & \textbf{0.376} & \textbf{0.125} & 0.135 & \textbf{0.184} & \textbf{0.471} & \textbf{0.532} & 1.377 \\
\hline
\end{tabular}
\label{tab:localization metrics}
\end{center}
\vspace{-10pt}
\end{table*}

Our results demonstrate that the choice of alignment method plays a crucial role in achieving sub-meter accurate localization estimation.
Table \ref{tab:localization metrics} shows that training without any alignment results in significant run-time localization error, while the vehicle-to-map approach outperforms the map-to-vehicle approach and provides RMS accuracy better than the 0.3m, 0.5$^{\circ}$ error threshold required for autonomous vehicle applications. 

An example localization result is shown in Fig. \ref{fig:map alignment on localization results}. 
We observe that different alignment approaches result in similar offline embeddings, but quite different online embeddings.
Improved alignment results in a more clearly-defined online embedding, since the model only needs to learn the modality gap and not the implicit alignment gap.

\subsection{Aerial Data Modalities}

\begin{figure}[t!]
\centering
\begin{subfigure}{0.31\columnwidth}
    \includegraphics[width=\textwidth]{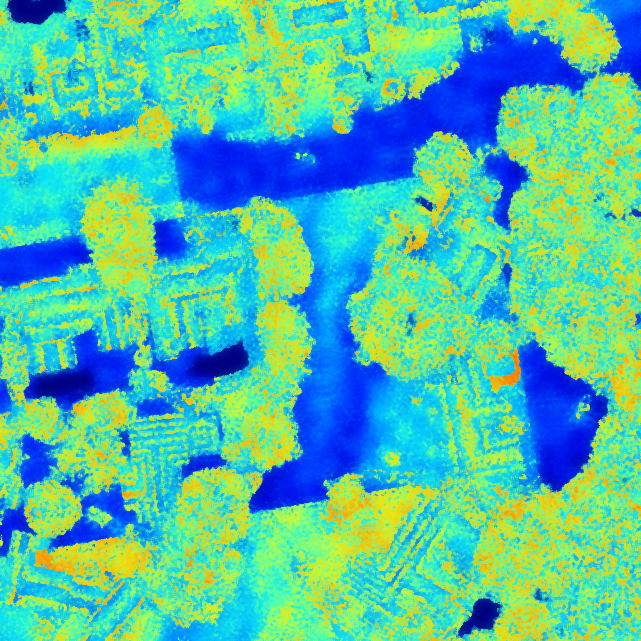} 
\end{subfigure}
\hfill
\begin{subfigure}{0.31\columnwidth}
    \includegraphics[width=\textwidth]{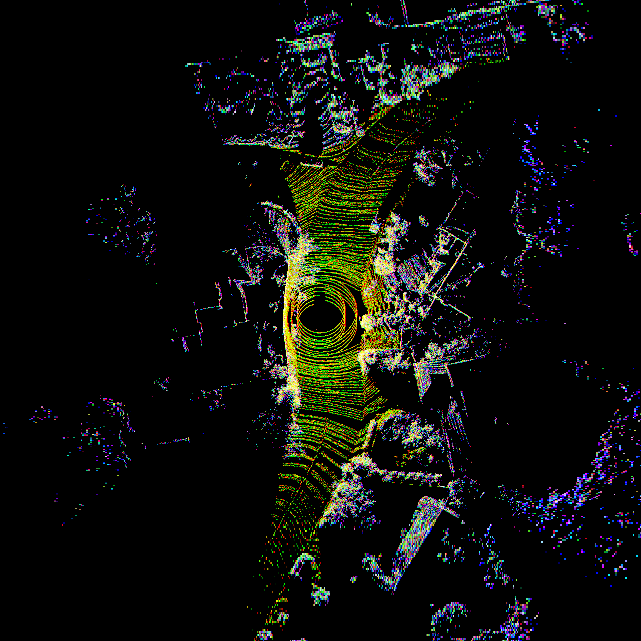}
\end{subfigure}
\hfill
\begin{subfigure}{0.31\columnwidth}
    \includegraphics[width=\textwidth]{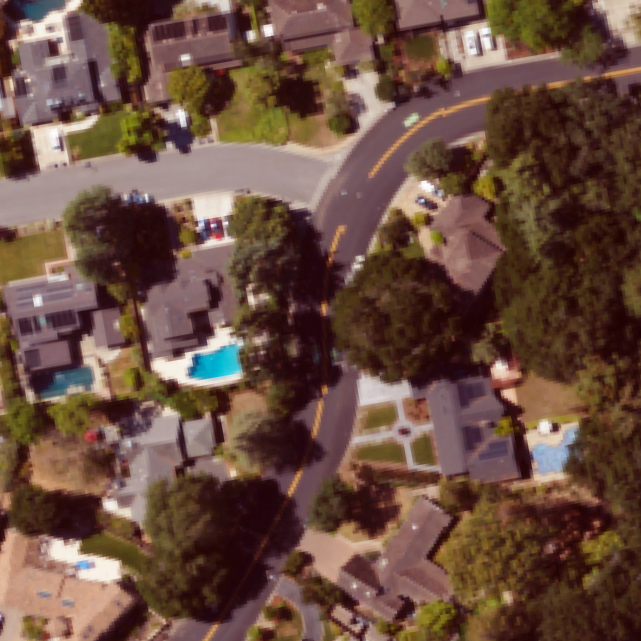}
\end{subfigure}

\vspace{2pt}

\begin{subfigure}{0.31\columnwidth}
    \includegraphics[width=\textwidth]{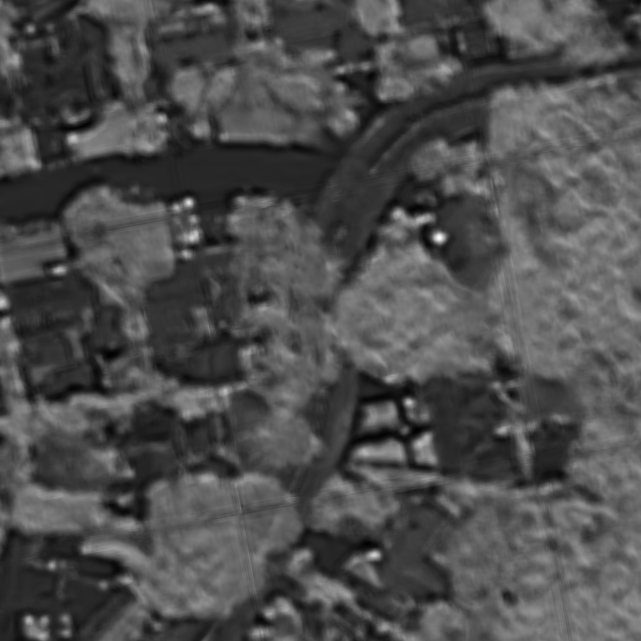}
\end{subfigure}
\hfill
\begin{subfigure}{0.31\columnwidth}
    \includegraphics[width=\textwidth]{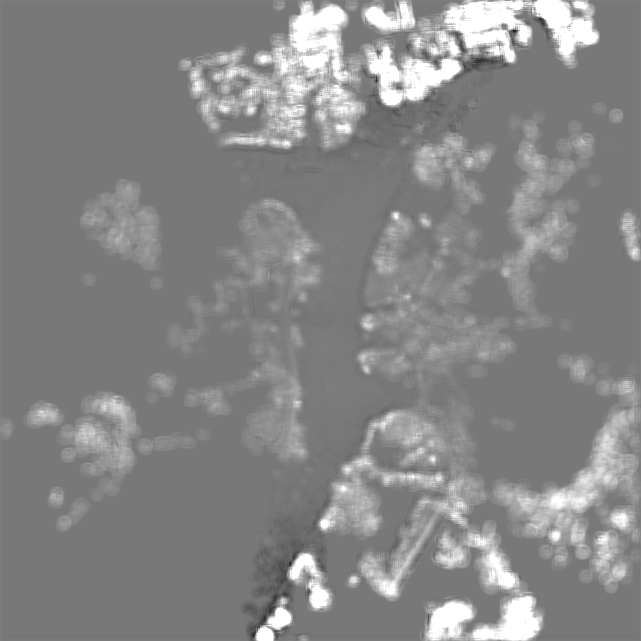}
\end{subfigure}
\hfill
\begin{subfigure}{0.31\columnwidth}
    \includegraphics[width=\textwidth] {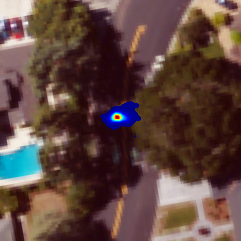}
\end{subfigure}

\vspace{2pt}

\begin{subfigure}{0.31\columnwidth}
    \includegraphics[width=\textwidth]{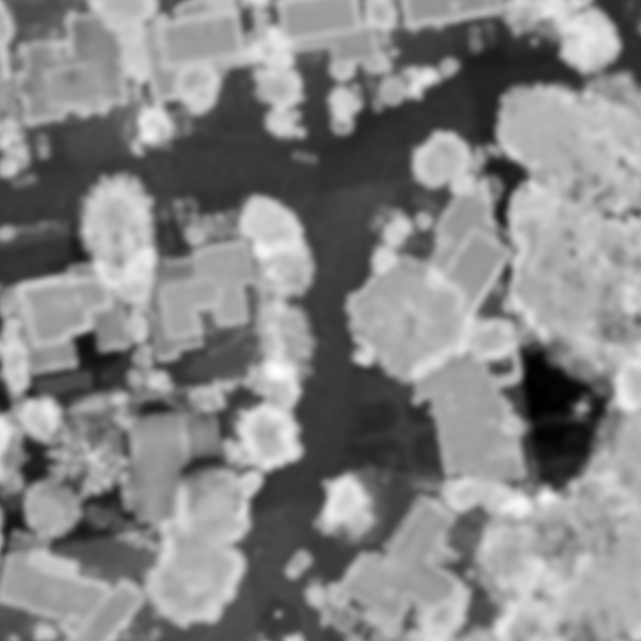}
\end{subfigure}
\hfill
\begin{subfigure}{0.31\columnwidth}
    \includegraphics[width=\textwidth]{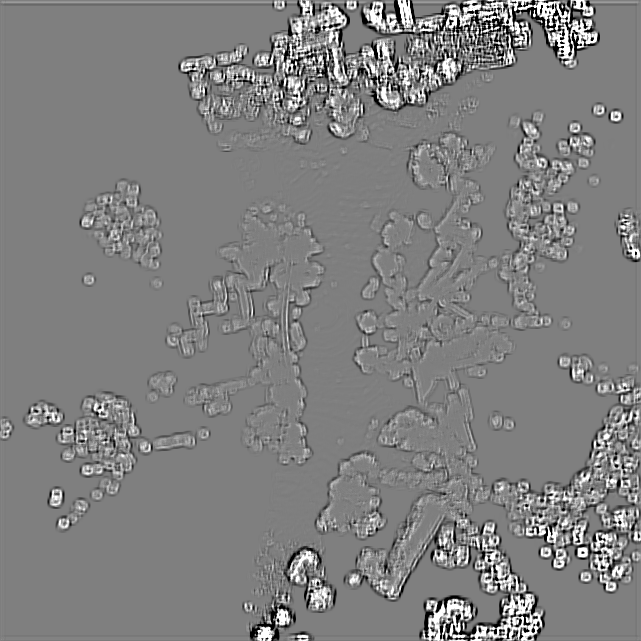}
\end{subfigure}
\hfill
\begin{subfigure}{0.31\columnwidth}
    \includegraphics[width=\textwidth]{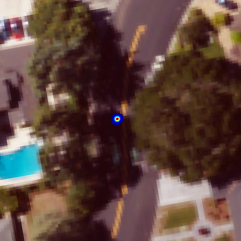}
\end{subfigure}

\caption{Localization results from models using either aerial RGB imagery only or aerial DSM only. The first row from left to right: USGS DSM, online LiDAR, USDA aerial imagery; subsequent rows show results from models trained with aerial RGB imagery only or aerial DSM data only respectively. 
}
\label{fig:different geo data on localization results}
\vspace{-10pt}
\end{figure}

Since the alignment approach is only required at training time, we can evaluate approaches that only have access to individual aerial data modalities at test time.
Fig. \ref{fig:different geo data on localization results} shows outputs from models trained with imagery geospatial data only or DSM geospatial data only on a single example.
Interestingly, a model trained with the vehicle-to-map alignment and deployed using only aerial RGB imagery (native resolution 0.6m / pixel) provides median errors below 0.3m and 0.5$^{\circ}$, paving the way for even lower cost aerial data sources for autonomous vehicle localization systems. 

\begin{figure}[t!]
\centering
\begin{subfigure}{0.31\columnwidth}
    \includegraphics[width=\textwidth]{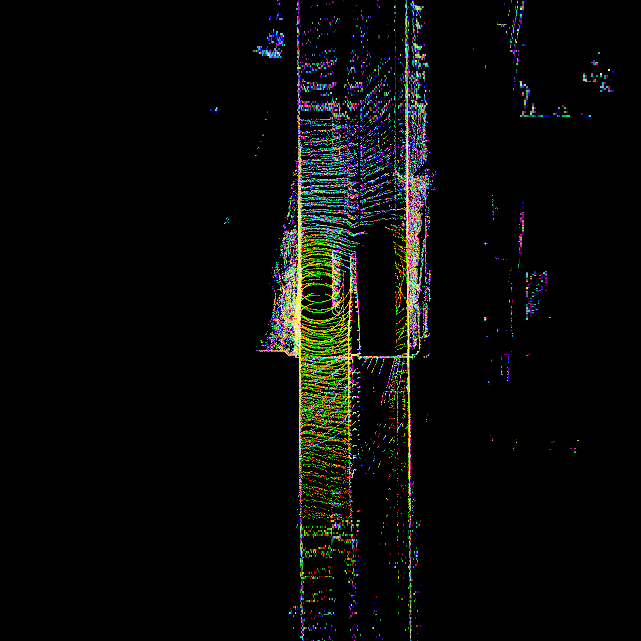}
\end{subfigure}
\hfill
\begin{subfigure}{0.31\columnwidth}
    \includegraphics[width=\textwidth]{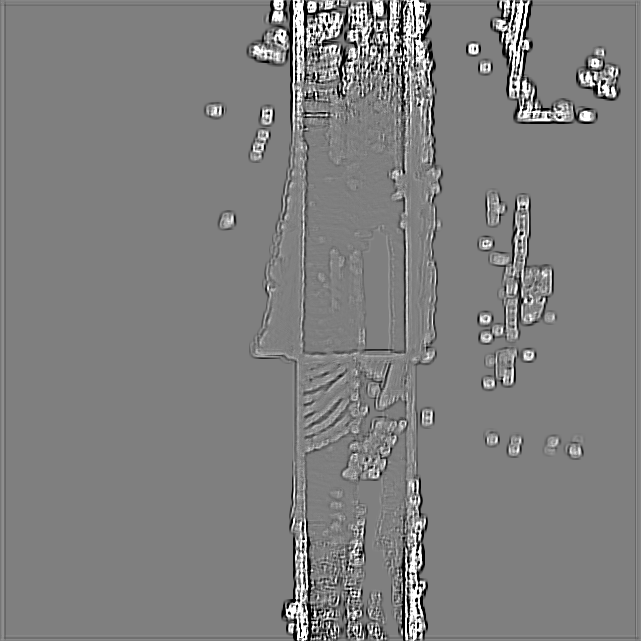}
\end{subfigure}
\hfill
\begin{subfigure}{0.31\columnwidth}
    \includegraphics[width=\textwidth]{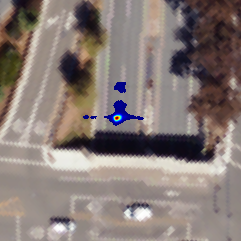}
\end{subfigure}

\vspace{2pt}

\begin{subfigure}{0.31\columnwidth}
    \includegraphics[width=\textwidth]{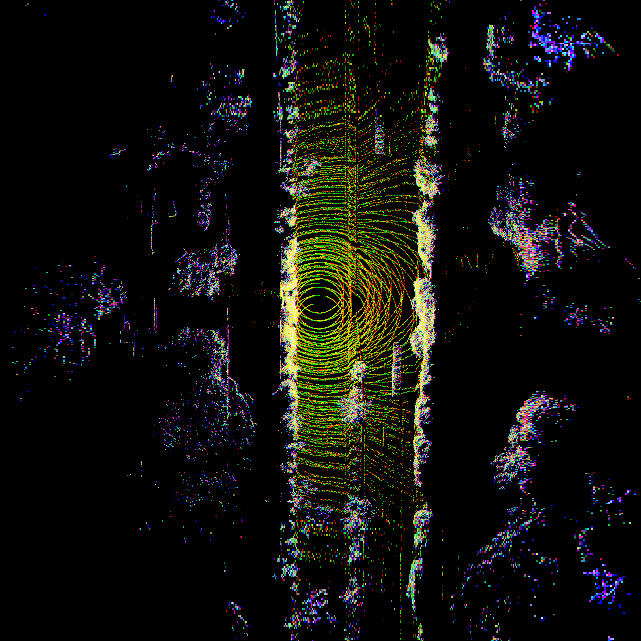}
\end{subfigure}
\hfill
\begin{subfigure}{0.31\columnwidth}
    \includegraphics[width=\textwidth]{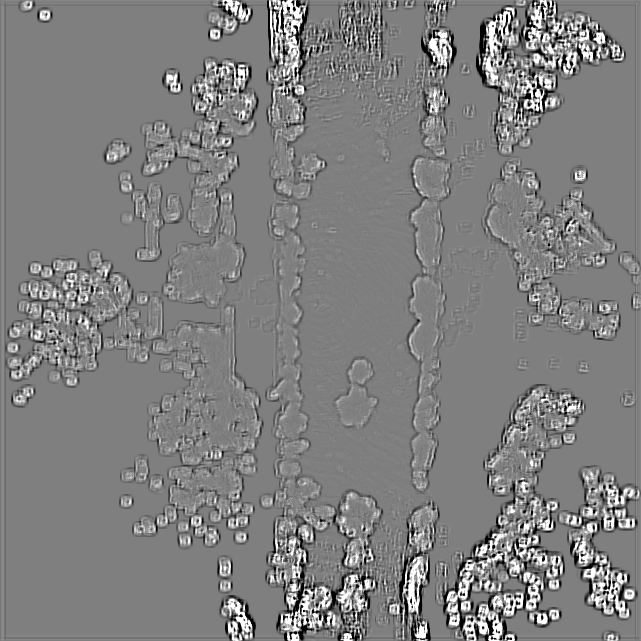}
\end{subfigure}
\hfill
\begin{subfigure}{0.31\columnwidth}
    \includegraphics[width=\textwidth]{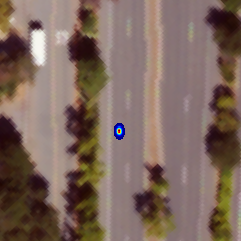}
\end{subfigure}

\caption{Some examples with relatively large localization errors. 
From left to right: online LiDAR, online embedding, USDA aerial imagery. In the top example, the vehicle is driving along an underpass that is not observable from the overhead imagery. 
In the lower example, the vehicle was driving along an arterial road which lacks longitudinal features. 
}
\label{fig:failure_modes}
\vspace{-10pt}
\end{figure}

\subsection{Failure Modes}
The P99 error column of Table \ref{tab:localization metrics} shows the methods still contain outliers upwards of 2m, and would need to be combined with a filtering or outlier rejection approach before direct use on an autonomous vehicle.
Some examples of failure modes are visualized in Fig. \ref{fig:failure_modes}. 
Common failure modes include insufficient longitudinal features, outdated aerial data, and occlusions between overhead aerial data and onboard perspective view.

%% file: sec/6_conclusion.tex
\section{Conclusion}
In this work, we identify that training data alignment is key to achieve sub-meter localization accuracy for learning-based localization using aerial data. 
We compare two different alignment methods to use jointly with a simple yet effective learned localization system using cross-correlation. 
Our large-scale experiments and ablation studies show that vehicle-to-map alignment outperforms tile-based map-to-vehicle alignment at training time, and leveraging aerial DSM and imagery data jointly can provide localization accuracy that exceeds the 0.3m, 0.5$^{\circ}$ guidance for autonomous vehicle application.
Even the method using only 0.6m / pixel resolution aerial RGB imagery to generate embeddings can provide a median accuracy better than 0.3m, 0.5$^{\circ}$ as a result of training-time alignment. 
Our results open a pathway to robust and precise localization with low-cost aerial data in the autonomous vehicle domain.